\documentclass[11pt,a4paper]{article}
\usepackage{times,latexsym}
\usepackage{url}
\usepackage[T1]{fontenc}
\usepackage{amsmath}
\usepackage{booktabs}
\usepackage{adjustbox}
\usepackage{tabularx}
\usepackage{tcolorbox}
\usepackage{etoolbox}
\usepackage{multirow}
\usepackage{algpseudocodex}
\usepackage{mdframed}
\usepackage{amsthm}
\usepackage{enumitem}
\usepackage{caption}

\usepackage[acceptedWithA]{tacl2021v1}

\usepackage{xspace,mfirstuc,tabulary}

\newif\iftaclinstructions
\taclinstructionsfalse 
\iftaclinstructions
\renewcommand{\confidential}{}
\renewcommand{\anonsubtext}{(No author info supplied here, for consistency with
TACL-submission anonymization requirements)}
\newcommand{\instr}
\fi

\iftaclpubformat 

\else

\fi

\usepackage{xcolor}
\usepackage{float}
\usepackage{placeins}

\newcommand{\dataset}{{\cal D}}

\usepackage{amsmath}
\usepackage{amsthm}
\usepackage{amssymb}
\usepackage{eqparbox}
\usepackage{mathtools}
\usepackage{scrhack}

\usepackage{booktabs}
\usepackage{colortbl}

\usepackage{hyperref}
\usepackage{cleveref}

\definecolor{tablecolor}{rgb}{0.8,0.8,0.8}

\newcommand\cut[1]{}

\newcommand{\squishlist}{
   \begin{list}{$\bullet$}
    { \setlength{\itemsep}{0pt}      \setlength{\parsep}{3pt}
      \setlength{\topsep}{3pt}       \setlength{\partopsep}{0pt}
      \setlength{\leftmargin}{1.5em} \setlength{\labelwidth}{1em}
      \setlength{\labelsep}{0.5em} } }

\newcommand{\squishlisttwo}{
   \begin{list}{$\bullet$}
    { \setlength{\itemsep}{0pt}    \setlength{\parsep}{0pt}
      \setlength{\topsep}{0pt}     \setlength{\partopsep}{0pt}
      \setlength{\leftmargin}{2em} \setlength{\labelwidth}{1.5em}
      \setlength{\labelsep}{0.5em} } }

\newcommand{\squishend}{
    \end{list}  }

{}
{}
{}

\newtheoremstyle{mydefstyle}
  {3pt}   
  {3pt}   
  {\itshape}      
  {}      
  {\bfseries} 
  {.}     
  {.5em}  
  {\thmname{#1}\thmnumber{ #2}\thmnote{ (\textbf{#3})}}

\theoremstyle{mydefstyle}

\newcommand{\myvec}[1]{\mbox{$\mathbf{#1}$}}

\newcommand{\vd}{\mbox{$\myvec{d}$}}

\newcommand{\vs}{\mbox{$\myvec{s}$}}

\newcommand{\vx}{\mbox{$\myvec{x}$}}

\newcommand{\vy}{\mbox{$\myvec{y}$}}

\newcommand{\calD}{\mbox{${\cal D}$}}

\newcommand{\calY}{\mbox{${\cal Y}$}}

\newcommand{\vocab}{\mathcal{V}}

\newcommand{\human}{\text{H}}
\newcommand{\ai}{\text{AI}}
\newcommand{\mixed}{\text{MX}}
\newcommand{\origin}{\text{au}}
\newcommand{\prompt}{\text{pr}}
\newcommand{\airate}{\text{ar}}

\tcbuselibrary{theorems}

\newtcbtheorem{definition2}{Definition}%
{%
  colback=gray!8,
  colframe=gray!60,
  boxrule=0.5pt,
  width=\linewidth,
  left=2pt, right=2pt,
  top=4pt, bottom=4pt,
  boxsep=2pt,
  arc=2pt,
  fonttitle=\bfseries,
  fontupper=\itshape,
  coltitle=black,
  title={Definition~\thetcbcounter%
    \ifstrempty{#2}{}{ (#2)}}
}{def}

\DeclareCaptionType[fileext=lon,placement=h]{notion}[Algorithm][List of Notions]
\algnewcommand\algorithmicnotion{\textbf{notion}}
\algblockdefx{Notion}{EndNotion}[1]
  {\algorithmicnotion\ #1}
  {}

\algrenewcommand{\algorithmicfunction}{\textbf{function}}
\algrenewcommand{\textproc}[1]{#1}

\title{'Your AI Text is not Mine': Redefining and Evaluating AI-generated Text Detection under Realistic Assumptions}

\author{
  Nils Dycke$^{1,2}$\thanks{\hspace{0.5em}These authors contributed equally to this work.} \quad
  Marina Sakharova$^{1,2,3}$\footnotemark[1] \quad
  Nico Daheim$^{1,2}$ \quad
  Iryna Gurevych$^{1,2,3}$ \quad \\
  $^{1}$Ubiquitous Knowledge Processing Lab (UKP Lab),\\Department of Computer Science, Technical University of Darmstadt\\
  $^{2}$National Research Center for Applied Cybersecurity ATHENE, Germany\\
  $^{3}$Zuse School ELIZA\\
}

\date{}

\algrenewcommand\algorithmicindent{1em}

\begin{document}
\maketitle

\begin{abstract} 
Although it is generally agreed that AI-generated text poses a broad societal risk, there is no common understanding in the AI-generated text detection literature on what constitutes harmful use. 
Rather, existing datasets and approaches often define their own criteria and make their own assumptions, sometimes implicitly, and often only loosely related to real-world needs and applications.
To address this gap, we here systematically define various notions of AI-generated text and their characteristics.
To study these, we collect AITDNA - a new benchmark of human-machine co-constructed texts that is annotated with detailed genesis information, such as the entire edit and AI-interaction history.
We benchmark various machine-generated text detectors and find that they often only perform well for specific notions but not as broad detectors.
We release code\footnote{\url{https://github.com/UKPLab/arxiv2026-aitdna}} and data\footnote{\url{https://huggingface.co/datasets/UKPLab/AITDNA}} publicly.
\end{abstract}

\section{Introduction} 
AI-generated text poses broad societal risks, including weakened quality control in science \cite{liang2024mapping}, bypassing university examinations \cite{gruenhagen2024rapid}, and large-scale misinformation campaigns \cite{barman2024dark}.
AI-generated text detection (AITD) is a key technology to mitigate these risks by identifying if a given text, or parts of it, were generated by a human or an AI \cite{gehrmann-etal-2019-gltr,mitchell2023detectgpt}. Although detection becomes increasingly challenging as Large Language Models (LLMs) assimilate to true human language use, reliable detection remains feasible \cite{pmlr-v235-chakraborty24a} and is therefore a key tool for filtering AI content. This sparked extensive research on the design and evaluation of AITD \cite{wu-etal-2025-survey}.

A broad range of task definitions has evolved. Many works formulate AITD as a document-level task, where a single author per text is assumed and a binary label is assigned to the entire document \cite{mitchell2023detectgpt, suDetectLLMLeveragingLog2023}. More recent studies investigate sentence-level detection \cite{jiang-etal-2025-sendetex,ijcai2024p835} and the identification of boundaries between human- and AI-generated passages \cite{dugan2023real,zeng2024towards}.
Although several benchmarks evaluate the cross-domain and cross-origin generalization of detectors to test their real-world robustness \cite[][i.a.]{dugan-etal-2024-raid,wu2024detectrl,wang-etal-2024-m4gt}, there is no systematic study of the \textit{underlying definitions}, i.e., the \textit{notions}, of AITD and their alignment with real-world use cases. This gap can be largely attributed to the lack of standardized language to specify and discuss AITD notions and gives rise to two key problems.

First, existing AITD notions in datasets and benchmarks are misaligned with real-world AI use and AI policies. Human-AI co-creation of texts has emerged as a common writing pattern across domains \cite{lee2022coauthor,mysore-etal-2025-prototypical}, where a human author and LLM collaboratively write text resulting in nuanced authorship specific to each token. Commonly, some form of co-creation is considered permissible in AI use policies \cite{wang2024generative,saha2026policies}. AITD notions and datasets need to explicitly account for human-AI co-writing to be able to differentiate benign from malicious AI use depending on the applicable policies. However, most prior work \cite[][i.a.]{wu2024detectrl,zhang-etal-2024-llm} does not frame AITD in terms of normative labels (permissible or prohibited) and relies on simplistic notion assumptions reflected in the synthetically constructed texts. This raises the question: do these definitions and their associated data truly reflect human-AI co-writing?

Second, existing AITD notions are commonly under-specified, making findings on different datasets incompatible. While some works consider AI-generated text as any text that was output by an LLM, authorship is often more nuanced since LLMs can reproduce genuinely human text, e.g., from their training data \cite{wuMembershipInferenceAttacks2025}, and paraphrase human input passages. This makes the assignment of authorship a complex problem \cite{triptoShipTheseusCurious2024}.
This complexity gives rise to a range of interpretations of the AITD problem, implying strong assumptions on the creation of the text and the downstream use of detectors, which are largely left unspecified. This raises the question: which assumptions are encoded in existing definitions and associated data, and are they compatible with real-world use cases?

In this paper, we rethink how to define and discuss AITD, as this is critical to provide policy makers with guidance on choosing the most appropriate notion and associated detectors for their use case and to make evaluation consistent and comparable across datasets. We address the existing gap with three key contributions.

\textbf{C\#1}: We analyze AITD notions across the literature and derive a framework for specifying AITD problems with explicit assumptions. We identify five notions and extend them by two new ones -- content-, and authorship-ID-based AITD -- to cover more real-world AITD use cases.

\textbf{C\#2}: To test existing AITD evaluation practices with respect to their hidden assumptions and alignment with human-AI co-writing, we collect AITDNA, a novel dataset of more than 350 texts from 99 human authors while recording detailed human-AI interactions during writing, covering multiple genres, LLMs, and generation temperatures. Unlike prior work, AITDNA includes the detailed prompts and text edits by the humans and the LLM under a realistic writing setting.

\textbf{C\#3}: We use AITDNA to check existing AITD datasets for their hidden assumptions and systematically study the role of the different AITD notions for evaluation. We find that existing AITD datasets do not represent natural human-AI co-creation and encode unrealistic assumptions. Controlling for notions, we systematically study which notion is the most difficult to detect, rank detectors across datasets by enforcing consistent assumptions, and estimate the effect of assumptions on performance evaluation empirically.

\section{Related Work}
\paragraph{Human-AI Co-writing in AITD}
Existing work on human-AI co-writing in AITD typically assumes simple forms of human-AI interaction as reflected in their synthetic datasets. DetectRL \cite{wu2024detectrl} and Mixset \cite{zhang-etal-2024-llm} assume human text is rewritten or polished by an LLM, while Sendetex \cite{jiang-etal-2025-sendetex} and \citet{zeng2024towards} assume AI-generated spans are inserted into human text. These datasets fail to reflect the interactive nature of co-writing \cite{mysore-etal-2025-prototypical}. Only CoAuthor \cite{lee2022coauthor}, as used by \citet{ijcai2024p835} for AITD, collects texts from human-AI interactions. However, CoAuthor focuses on short-text continuation without user prompts, neglecting text revision prompts, and, based on only one LLM, limiting its use for AITD evaluation. Our dataset, AITDNA, fills this gap by collecting the first human-AI co-writing dataset with detailed interactions including prompts and text revision requests with multiple LLMs.

\paragraph{Real-world alignment of AITD}
Prior work on alignment of AITD with real-world needs focuses on theoretical questions of authorship \cite{triptoShipTheseusCurious2024} or tests detectors on different domains and LLMs \cite{dugan-etal-2024-raid,wang-etal-2024-m4gt}. Unlike our study, these works do not consider different notions and assumptions across settings but only test different detectors. Most similar to our work, \citet{zhang-etal-2024-llm} and \citet{saha-feizi-2025-almost} test detectors on LLM-polished text in alignment with many AI policies; \citet{saha2026policies} test this specifically for peer reviews. However, these works assume a simplified human-AI text interaction paradigm and do not systematically compare different notions. Finally, \citet{wang-etal-2024-m4gt} evaluate detectors on a benchmark with varying detection granularity, but rely on synthetic data and cover only a small set of notions.

\section{AITD Notions} \label{sec:notions}
Task definitions of AITD, \textit{notions}, are diverse and often rely on hidden assumptions encoded in the associated evaluation data and detectors. We establish a shared vocabulary to organize notions.

\subsection{Basics}
Natural language text is typically treated as a static data point, but humans actually produce texts iteratively by creating, revising, and deleting tokens \cite{kuznetsov2022revise}. This creation process, or \textit{genesis}, usually goes unrecorded, reducing the sequence of \textit{text interactions} to a plain string. In AITD, these interactions are a key signal to defining authorship by the edits from the human author and the AI.
Formally, let $\vx \in  \dataset$ denote a text with tokens $\vx = (x_1, ..., x_{T})$ with $x_t \in \vocab$, a fixed vocabulary. We refer to the subsequence $(x_i, ..., x_t)$ with $i < t$ as the \textit{span} $\vx_{i:t}$. \Cref{tab:notation} provides an overview of the notation and the symbols used.

\begin{definition2}{Genesis}{}
    The genesis $\text{gen}=(\origin, \prompt)$ of $\vd$ is given by $\origin$ assigning each token a label from $\calY_{\text{gen}}=\{\human, \ai, \mixed\}$ representing human, AI, and human-AI-mixed origin (e.g. AI-corrected human-originated), respectively. For each token $x$ with $\origin(x) \in \{\ai, \mixed\}$, $\prompt(x)$ associates the prompt underlying the generation.
\end{definition2}

\noindent A text’s genesis is an objective property, describing \textit{what is}, not \textit{what should be}. In contrast, AITD notions define detection goals based on assumptions and normative standards for AI use, often referencing the text’s genesis. These standards distinguish malicious from benign use, yielding different detection labels for the same texts to train or evaluate detectors.

\begin{table}[t]
    \centering
    \small
    \begin{tabular}{ll}
      \toprule
      \textbf{Symbol} & \textbf{Meaning} \\
      \midrule
      $\calD$ & Dataset of texts \\
      $\vocab$ & Vocabulary of the LLM \\      
      $x$ & A single token \\
      $\vx$ & Text segment \\
      $\vx_{i:t}$ & Text from token $i$ to $t$ \\
      $\vd$ & The entire document \\
      $T$ & Total number of tokens in $\vd$ \\
      $\calY$ & Label space for detection \\
      $y$ & A single label  \\
      $\vy$ & Label sequence for a span \\
      $\prompt(\vx)$ & Prompt generating $\vx$ \\
      $\origin({\vx})$ & Author of $\vx$ \\
        $\airate(\vx)$ & AI token ratio \\
        $\tau$ & AI ratio threshold \\
        $\beta$ & Number of boundaries \\
      \bottomrule
    \end{tabular}
    \caption{Overview of the used notation.}
    \label{tab:notation}
  \end{table}

\subsection{Structuring AITD Notions}
\begin{figure*}[t]
    \centering
    \includegraphics[width=0.8\linewidth]{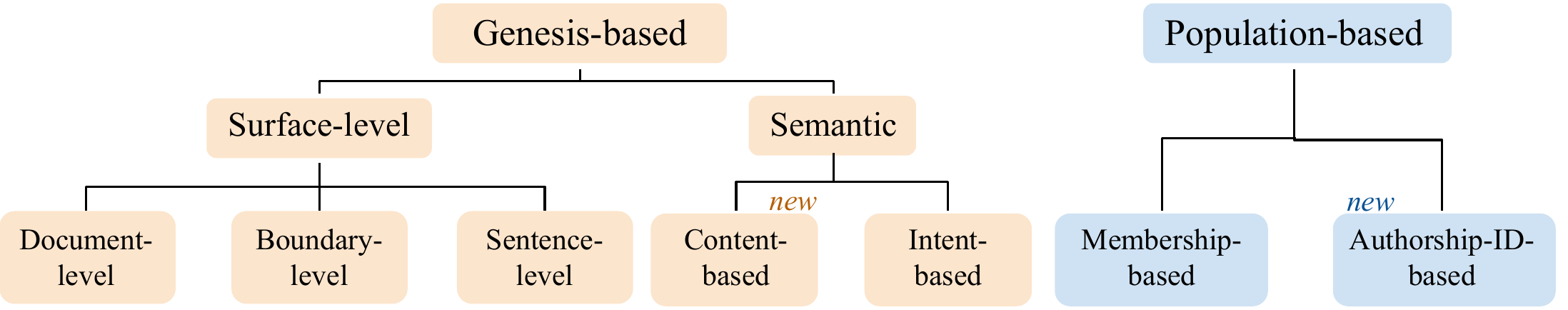}
    \caption{Structuring AITD notions (detailed survey results in \Cref{tab:notions}).}
    \label{fig:taxonomy}
\end{figure*}

We analyze AITD notions along three dimensions:
(1) the \textbf{normative standard} distinguishes acceptable from unacceptable AI use; for example, AI may be entirely prohibited, or allowed only if most text comes from a human; (2) the \textbf{genesis assumptions} impose constraints on a text’s genesis to simplify detection; for instance, a text may be assumed to be produced \textit{en-bloc} by either an AI or a human; (3) the \textbf{attacker model} indicates the degree to which the genesis assumptions are violated, framed as an approach to blend unacceptable into acceptable AI use; for example, sentences in a fully AI-generated text may be replaced with human-written ones, violating the en-bloc assumption partially.
Using these criteria, we identify five common AITD notions from the literature and define two new ones to fill practical gaps, as shown in \Cref{fig:taxonomy} (see \Cref{tab:notions} for a full overview). The notions fall into two groups: \textbf{genesis-based} notions, which define normative standards based on text genesis and ask \textit{did a human produce this span?}, and \textbf{population-based} notions,  relying on a reference population and ask \textit{does this span resemble human text?}

\paragraph{Genesis-based notions}
Genesis-based notions are most prominent. They differ in genesis assumption granularity and, thus, detection targets. Specifically, \textbf{document-level AITD} \cite[e.g.][]{wu2024detectrl} assumes en-bloc generation by a human or an AI, \textbf{boundary-level AITD} \cite[e.g.][]{zeng2024towards} assumes a fixed number of alternating passages between human and AI, and \textbf{sentence-level AITD} \cite[e.g.][]{jiang-etal-2025-sendetex} assumes human- and AI-generated sentences. All three notions adopt a similar attacker model, in which selected tokens are replaced post-hoc.
These assumptions imply distinct normative standards: at the document level, no AI use is benign, and a detector should flag any text containing a certain amount of AI-generated content, although this threshold is often unspecified. In contrast, on boundary- or sentence-level, detection supports a nuanced view of AI use, as detectors identify AI-generated spans at a fine-grained level, leaving the final decision about the document to the user.
These notions define normative standards independently of the \textit{content} or \textit{intent} of AI passages. However, policies for AI use typically focus on these aspects. To capture such \textit{semantic} goals, we define \textbf{intent-based AITD}, aligning with \citet{zhang-etal-2024-llm}, as a variant of sentence-level AITD, which flags AI-generated sentences whose underlying prompts violate an intent policy, and newly introduce \textbf{content-based AITD}, which flags AI-generated sentences that violate a content policy. 

\paragraph{Population-based notions}
The dominant population-based notion is \textbf{membership-based AITD}, advocated in \cite{triptoShipTheseusCurious2024,koike2025machine}.  Its key idea is that AI can generate genuinely human text, for example, by reproducing training data or common phrases. 
Therefore, detection does not depend on the true author of a span but on its membership in a reference corpus. In this view, some AI use is acceptable, as in sentence-level AITD, and only spans outside the population are flagged.
This notion can be adapted to better align with real-world needs by varying the reference population. We propose \textbf{authorship-identification-based AITD}, which uses a reference corpus specific to a given human author rather than a general human population. This formulation effectively casts AITD as an authorship identification task.

\subsection{Standardization}
We describe notions by their text segmentation and label assignment jointly encoding the normative standard, genesis assumptions, and attacker model.

\begin{definition2}{Notion}{}
    A notion $N=(B, M)$ maps each token $x$ in $\vd$ to a label $l\in \calY = \{\human, \ai\}$, where
    $B$ is a partition algorithm splitting $\vd$ into $k$ contiguous, non-overlapping, and covering spans $B(\vd)=\{\vx_1, ..., \vx_k\}, k\in\{1,...,T\}$.
    $M: \vx \mapsto \calY$ assigns a detection label to each span $\vx \in B(\vd)$. 
\end{definition2}

Genesis-based notions segment and label text based on the genesis $\text{gen}$. We gauge the normative standard and attacker model in the parameter $\tau \in \left[0, 1\right]$ defining the degree to which an assigned $\ai$ label of a segment may violate its underlying genesis. Formally, we define $M_\text{gen}$ using the ratio of AI tokens of a span as given by the genesis $\text{ar}(\vx_{i:j})=\frac{|\{k | k \in \left[i, j\right] \text{ and } \origin(x_{k}) = \ai\}|}{j-i+1}$:
$$
    M_\text{gen}^\tau(\vx_{i:j}) = 
    \begin{cases}
        \ai \text{, if } \text{ar}(\vx_{i:j}) \geq \tau \\
        \human \text{, else}
    \end{cases}
$$

\noindent Document-level AITD segments the text into a single span that covers the entire document and assigns labels by $M_\text{gen}^\tau$.

\begin{definition2}{Document-level AITD}{}
    Document-level AITD is a notion \mbox{$\displaystyle N_\text{doc}^\tau=(B_\text{doc}, M_\text{gen}^\tau)$} with
    $B_\text{doc}(\vd) = \{\vx_{1:T}\}$
\end{definition2}

\noindent Sentence-level AITD segments the document into sentences and assigns labels by $M_\text{gen}^\tau$.

\begin{definition2}{Sentence-level AITD}{}
    Sentence-level AITD is a notion \mbox{$\displaystyle N_\text{sen}^\tau=(B_\text{sen}, M_\text{gen}^\tau)$} with
    $B_\text{sen}(\vd) = \{\vx_1, ..., \vx_k\}$ where  $\vx_i$ is a sentence.
\end{definition2}

\noindent Boundary-level AITD assumes exactly $\beta$ boundaries between alternating human and AI passages, placed to agree maximally with the genesis. Practically, these boundaries are found via dynamic programming with the following objective.

\begin{definition2}{Boundary-level AITD}{}
    Boundary-level AITD is a notion
    $N_\text{bou}^\beta=(B_\text{bou}^\beta,M_\text{bou})$ with
    $\beta\in\mathbb{Z}^+$. Let $\mathcal{Y}_\beta\subseteq\calY^T$ be the
    label sequences $\vy=(y_1,\dots,y_T)$ for $\vd$ that switch between
    $\human$ and $\ai$ exactly $\beta$ times, and let $\vy^{*}$ be the one
    that disagrees with the genesis on the fewest tokens,
    $$
        \vy^{*}=\arg\min_{\vy\in\mathcal{Y}_\beta}
        \big|\{\,x_t : y_t\neq\origin(x_t)\,\}\big|.
    $$
    Then $B_\text{bou}^\beta(\vd)$ splits $\vd$ at the $\beta$ switches of
    $\vy^{*}$, and $M_\text{bou}(\vs)=y^{*}_t$ for any token $x_t\in\vs$.
\end{definition2}

\noindent This makes $\tau$ and $\beta$ key parameters of genesis-based notions warranting explicit specification.

Content-based and intent-based notions employ sentence-level segmentation but sub-select AI-generated sentences according to a policy \mbox{$\displaystyle \rho: \cdot \mapsto \mathbb{B}$} with the semantic labeling function:
$$
M_{\text{sem}}^{\rho, \tau}(\vx) = 
    \begin{cases}
    	\ai \text{, if } M_\text{gen}^\tau(\vx) = \ai \land \rho(\vs) = \bot \\
    	\human \text{, else}
    \end{cases}
$$
\noindent For content-based AITD, the policy considers only the span's content. 
\begin{definition2}{Content-based AITD}{}
    Content-based AITD is a notion \mbox{$\displaystyle N_\text{cnt}^{\rho_\text{cnt},\tau}=(B_\text{sen}, M_{\text{sem}}^{\rho_\text{cnt}, \tau})$} with a content policy \mbox{$\displaystyle \rho_\text{cnt}: \vx_{i:j} \mapsto \mathbb{B}$}.
\end{definition2}

\noindent For intent-based AITD, the policy considers only its underlying prompt.
\begin{definition2}{Intent-based AITD}{}
    Intent-based AITD is a notion \mbox{$\displaystyle N_\text{int}^{\rho_\text{int},\tau}=(B_\text{sen}, M_{\text{sem}}^{\rho_\text{int}, \tau})$} with an intent policy \mbox{$\displaystyle \rho_\text{int}: \prompt(\vx_{i:j}) \mapsto \mathbb{B}$}.
\end{definition2}

\noindent Population-based notions are parameterized by a reference population $\mathcal{P}$. Here, segmentation is intertwined with labeling, as it relies on matching $\text{n-grams}(\vs)$ of a span with $n \geq 1$ to $\mathcal{P}$. Practically, this splits the text into spans that consist of matching sequences of n-grams and those that do not.
\begin{gather*}
B_{\mathcal{P}}^+(\vd) = \{
\vs \in \vd | \text{ n-grams}(\vs) \subseteq \mathcal{P};
\vs \text{ is maximal}
\} \\
B_{\mathcal{P}}^-(\vd) = \{
\vs \in \vd | \text{ } \vs \text{ not covered by } B_{\mathcal{P}}^+(\vd)
\}
\end{gather*}

\noindent Then, segmentation and labeling are realized via $B_{\mathcal{P}}(\vd) = B_{\mathcal{P}}^+(\vd) \cup B_{\mathcal{P}}^-(\vd)$ and
$$
M_\mathcal{P}(\vs)  = 
    \begin{cases}
    	\human \text{, if } \vs \in B_{\mathcal{P}}^+(\vd) \\
    	\ai \text{, else}
    \end{cases}
$$

\noindent Membership-based AITD uses a population describing the canon of human text, whereas authorship-ID-based AITD filters such a population by the specific author of the document.

\begin{definition2}{Membership-based AITD}{}
    Membership-based AITD is a notion $N_\text{mem}^{\mathcal{P}, n} =(B_\mathcal{P}, M_{\mathcal{P}},n)$.
\end{definition2}

\begin{definition2}{Authorship-ID-based AITD}{}
    Authorship-ID-based AITD is a notion $N_\text{aid}^{\mathcal{P}, n}=(B_{f_{\text{\vd}}(\mathcal{P})}, M_{f_{\text{\vd}}(\mathcal{P})},n)$ where $f_{\text{\vd}}(\mathcal{P})$ returns documents of the author of $\vd$.
\end{definition2}

\subsection{Takeaways}
We make two key observations.
First, existing notions rely on distinct assumptions, many of which are under-specified. As a result, the same text with identical genesis may be considered acceptable under one notion but not another. Even within a single notion, parameters reflecting different normative standards can produce different labels. The alignment between detectors and these different target labels warrants empirical study.
Second, each notion is designed for a specific use case in which its assumptions are reasonable (\Cref{tab:notions}). For example, document-level AITD is suitable when aiming to detect AI-generated propaganda or bot networks. Applying a notion outside its intended context may violate its assumptions, making it essential to verify them before use for evaluation or training.
%
%

\section{Notions and AITD Datasets}
The above analysis shows that notions of AITD vary across the literature. Existing datasets encode these notions using synthetic data generation and heterogeneous data sources, while often leaving their assumptions under-specified. This limits direct comparison across datasets. We introduce AITDNA, a dataset collected without targeting a specific notion. It is therefore \textit{notion-agnostic} and aligned with real-world human–AI co-creation. AITDNA enables controlled comparison of notions, cross-notion evaluation, and analysis of assumptions in prior datasets.

\subsection{The AITDNA Dataset}
To construct AITDNA notion-agnostic and close to a natural human–AI co-creation, we design a dedicated co-writing interface based on CARE \cite{zyska-etal-2023-care}. The interface keeps detailed logs of a text’s genesis and captures typical modes of human–AI co-creation.

\paragraph{Interface}
Participants co-create with an LLM in a text editor. They write text directly and request LLM support in two modes: \textit{continuation} and \textit{revision}. In continuation, the LLM extends the text from a selected position, optionally guided by a prompt. In revision, the user selects a span, provides an optional prompt, and the LLM edits the span. In both modes, the output displays next to the editor, for the user to accept or reject the edit.\footnote{Appendix \Cref{fig:care-interface} shows the interface.}

Besides the final text, the interface records all text edits by the participant and the LLM, resulting in \textit{interaction traces} covering detailed insertions and deletions, user prompts, and LLM responses over time. Based on these fine-grained traces, we derive the authorship and underlying prompts for each token, yielding the text's genesis.

\paragraph{Setup}

\begin{figure}
    \centering
    \includegraphics[width=0.9\linewidth]{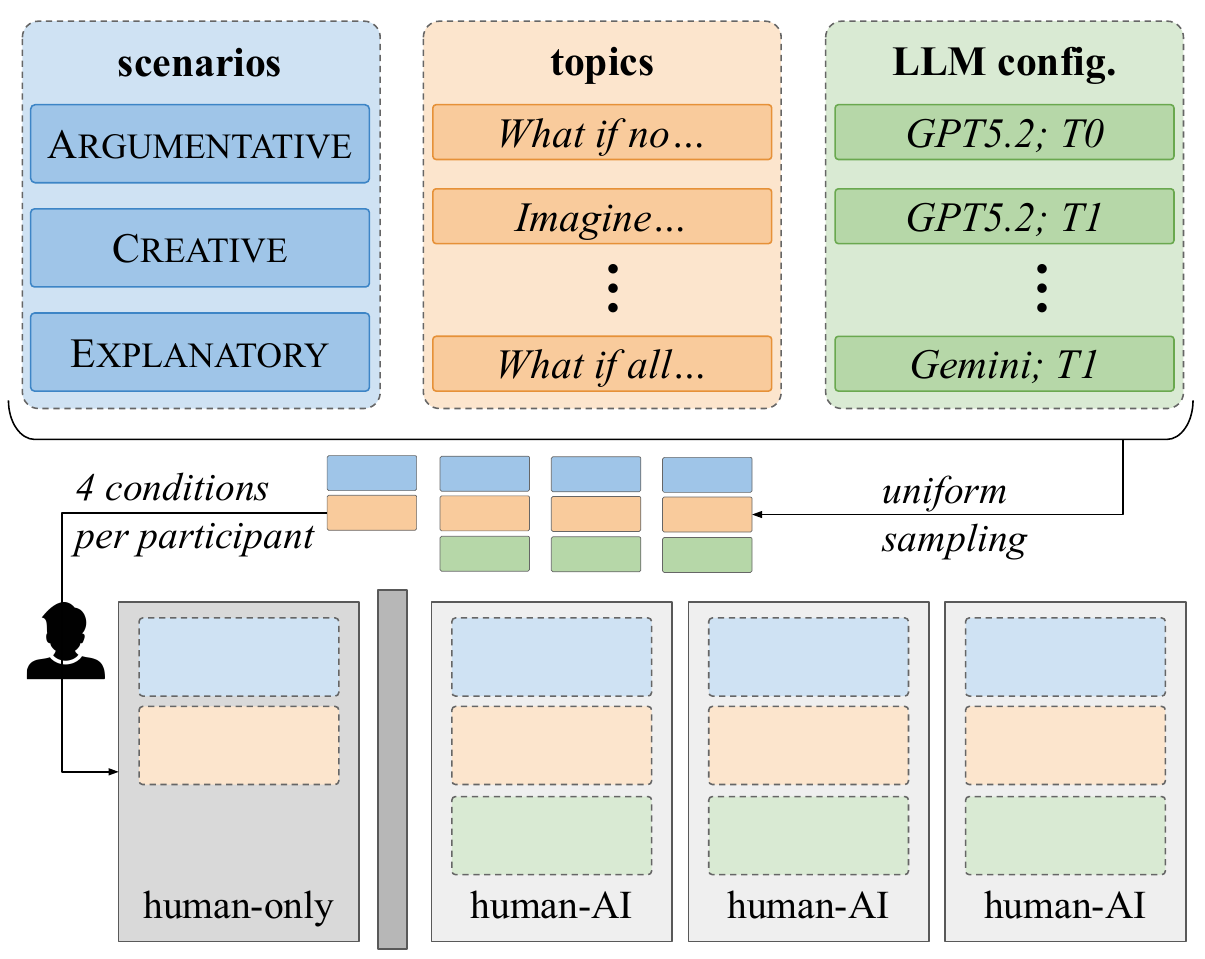}
    \caption{Overview of the study setup. We sample scenarios, topics, and LLM configurations and assign them in random order to the four conditions.}
    \label{fig:study-setup}
\end{figure}

We collect data across multiple sessions in a consistent setup illustrated in \Cref{fig:study-setup}. In each session, participants write in four conditions within roughly 90 minutes, targeting 350–450 words per text, similar to CoAuthor. Sessions take place in person or virtually, with one co-author providing instructions and answering questions.

In the first condition, participants write without LLM support.\footnote{Using the same interface with LLM support turned off.} This yields a subset of purely human texts necessary for population-based approaches and familiarizes participants with the setup. In the subsequent three conditions, participants co-create three texts with an LLM.
For each participant, we randomly assign a writing task per condition from three scenarios: \textsc{argumentative} (debating a topic), \textsc{creative} (writing a creative essay), and \textsc{explanatory} (explaining a topic). For \textsc{argumentative} and \textsc{creative}, we present two randomly sampled topics from the New York Times Student Discussions\footnote{\url{https://www.nytimes.com/column/learning-student-opinion}} from which participants choose one. For \textsc{explanatory}, participants define their own topic to ensure expertise. We enforce diverse sampling and avoid topic overlap across conditions.
For all co-creation conditions, we assign an LLM configuration by uniformly sampling from combinations of three open-weight models (Llama4-Scout \cite{meta2025llama4scout}, Qwen2.5:7B \cite{qwen2025qwen25technicalreport}, DeepSeek-V3.2 (Non-Thinking Mode), \cite{deepseekai2025deepseekv3technicalreport}) and two proprietary models (GPT-5.2 \cite{singh2026openaigpt5card}, Gemini-3-flash \cite{google2025gemini3flash}), with sampling temperatures $0$ or $1$.
In addition, we complement the essay-style scenarios with two sessions focused on scientific writing. Participants take a stance on a scientific topic (\textsc{argumentative}), develop an impact statement (\textsc{creative}), explain a research concept (\textsc{explanatory}), and perform a peer review.
Overall, this setup yields diverse topics and LLM configurations randomized across participants, suitable for a reliable AITD evaluation dataset \cite{wu-etal-2025-survey}.

\paragraph{Data collection}
We recruit participants in a university context, via the crowdsourcing platform Prolific\footnote{\url{https://app.prolific.com/}}, and from voluntary participants in writing workshops. All paid participants receive at least minimum-wage compensation. In total, 99 participants contribute texts across 8 sessions, including 2 sessions on scientific writing. All participants have at least B1-level English proficiency. For the scientific scenarios, we employ postgraduate or PhD students. For crowd workers, we select for high essay-writing performance by filtering participants in a pilot round.

Although AITDNA targets naturally produced text, quality control is required to remove data points with unmarked AI-generated text copied from outside the interface. We discard texts with copy–paste events involving spans above 60 characters from external sources based on the interaction traces; this heuristic is overly selective but ensures high data quality.

\paragraph{Resulting data} 
We collect 452 texts, with 362 remaining after filtering. 71\% of texts are written in essay style and 29\% in the scientific setting. Of all texts, 74\% are human-AI co-written; 95 samples are written in the human-only condition. Despite post-hoc filtering, all three writing scenarios are roughly equally represented, with an average of 117 texts per scenario, except for peer review with 10 texts. 
The texts have 459 words tokenized by whitespaces on average. While AITDNA is comparatively small, limiting the training of detectors, we deem it sufficient for analysis and evaluation, in line with recent human-created datasets in other domains \cite[e.g.][]{byun-etal-2026-cradle}.

\subsection{Hidden Assumptions in AITD Datasets} \label{ssec:assumptions-in-datasets}

\begin{table}[t]
    \centering
    \small
    \begin{tabular}{rccc}
    \toprule
        \textbf{Dataset} & \textbf{\%AI tokens} & \textbf{\# Boundaries} \\
        \midrule
        \textbf{AITDNA}   & 50.91\%   & 23.33 \\
        \textbf{CoAuthor}  & 24.09\%  & 20.32 \\
        \textbf{SenDetEx}  & 30.96\% & 13.57  \\
        \textbf{BD}        & 65.68\% & 2.75  \\
        \textbf{Mixset}    & 64.57\% & 31.51 \\
        \textbf{DetectRL}  & 63.69\% & 50.33 \\
        \bottomrule
    \end{tabular}
    \caption{Mixing of human and AI text, averaged across texts per dataset and metric.}
    \label{table:data-metric-compar}
\end{table}

\begin{figure}[t]
    \centering
    \includegraphics[width=0.95\linewidth]{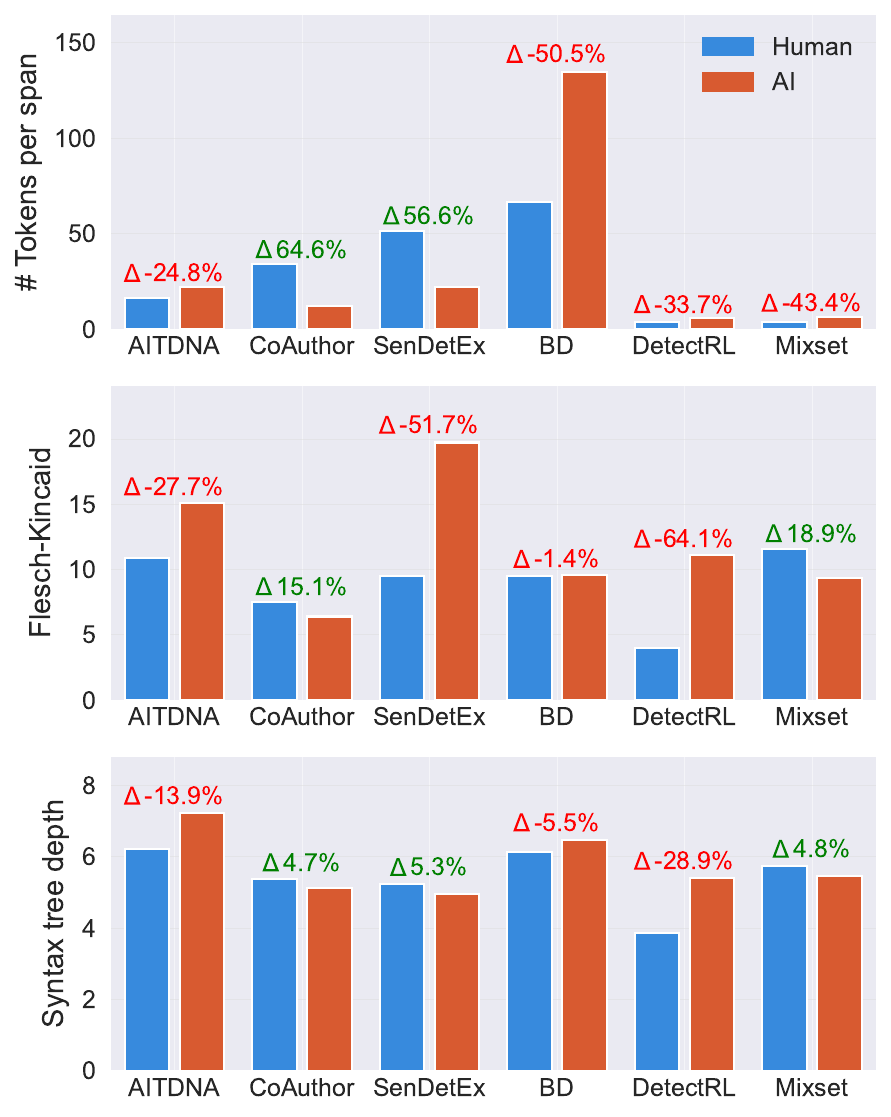}
    \caption{Language characteristics of human (blue) and AI (red) text spans per dataset. $\Delta$ reports the normalized difference between $\human$ and $\ai$.}
    \label{fig:ai-human-diffs}
\end{figure}

Except for CoAuthor, all prior datasets rely on notion-specific synthetic data generation, only mimicking human-AI co-creation. This raises the question: how well-aligned are the assumptions of these datasets to natural co-creation?

We analyze five widely used datasets based on genesis-based notions, as population-based notions have been introduced conceptually but lack readily available datasets. We consider CoAuthor \cite{lee2022coauthor} as a notion-agnostic dataset as used by \citet{ijcai2024p835} for sentence-level AITD; SenDetEx \cite{jiang-etal-2025-sendetex} as a sentence-level dataset; BD \cite{zeng2024towards} as a boundary-level dataset; and Mixset \cite{zhang-etal-2024-llm} together with the \textit{Polish Using LLM} portion of DetectRL \cite{wu2024detectrl} as document-level datasets. 
To construct SenDetEx, we use the original codebase\footnote{We use the code at \url{https://github.com/TristoneJiang/SenDetEX}, but correct for a misalignment with the method described in the paper and generate each text iteratively up to 10 times until its perplexity lies below that of the original.} and hyperparameters. To construct Mixset and BD, we infer the underlying genesis per text for each dataset. We start with the human-written text and reproduce AI edits through sequence alignment with Diff Match Patch\footnote{\url{https://github.com/google/diff-match-patch}}, which trades off semantic and syntax alignment. Finally, since AITDNA reflects natural modes of human-AI co-creation, we use it as a neutral point of reference to analyze alignment of prior datasets.

\paragraph{Genesis assumptions}
We turn to the genesis in AITD datasets to analyze how AI contributes to their texts quantitatively and linguistically.
\Cref{table:data-metric-compar} reports the average ratio of AI-generated tokens per text and the number of boundaries between AI and human spans, both measuring the degree of intermixing. \Cref{fig:ai-human-diffs} compares human and AI spans in terms of token count, Flesch-Kincaid grade level \cite{Kincaid1975DerivationON}, and syntax tree depth to assess linguistic similarity.

CoAuthor and SenDetEx have the lowest proportion of AI tokens. AITDNA lies near 50\%. BD, Mixset, and DetectRL lie at the top with around 65\%. This reflects substantial differences in the genesis assumptions, with Mixset and DetectRL clearly encoding a notion-typical \textit{en-bloc} genesis, where a fully human-written text is rewritten by AI into a completely new version in a single step.
As \Cref{fig:ai-human-diffs} shows, BD contains by far the longest continuous human and AI spans, resulting in few boundaries and low overall mixing (low \# Boundaries in \Cref{table:data-metric-compar}). In contrast, DetectRL and Mixset contain the shortest spans and boundaries, indicating strong mixing.

Regarding language complexity (\Cref{fig:ai-human-diffs}), CoAuthor and Mixset are the only datasets in which human text is more complex than AI text. DetectRL shows the largest difference between human and AI spans, likely due to its simplistic perturbation strategy mimicking human-AI co-creation. In SenDetEx, AI spans contain complex terminology (high Flesch-Kincaid) but simple syntax. AITDNA generally exhibits high language complexity, likely due to the scientific portion, while the differences between human and AI spans lie between the other datasets.

Overall, all datasets differ substantially from AITDNA in how human and AI text combine. This shows that synthetic text generation misaligns with natural human-AI co-creation. Although CoAuthor, like AITDNA, was collected without a target notion, its constrained co-writing setup, based on single-sentence continuation, produces longer spans overall, particularly for human text. These patterns differ notably from AITDNA and highlight the importance of careful collection design of human-AI co-writing datasets for AITD.

\begin{figure}[t]
    \centering
    \includegraphics[width=0.95\linewidth]{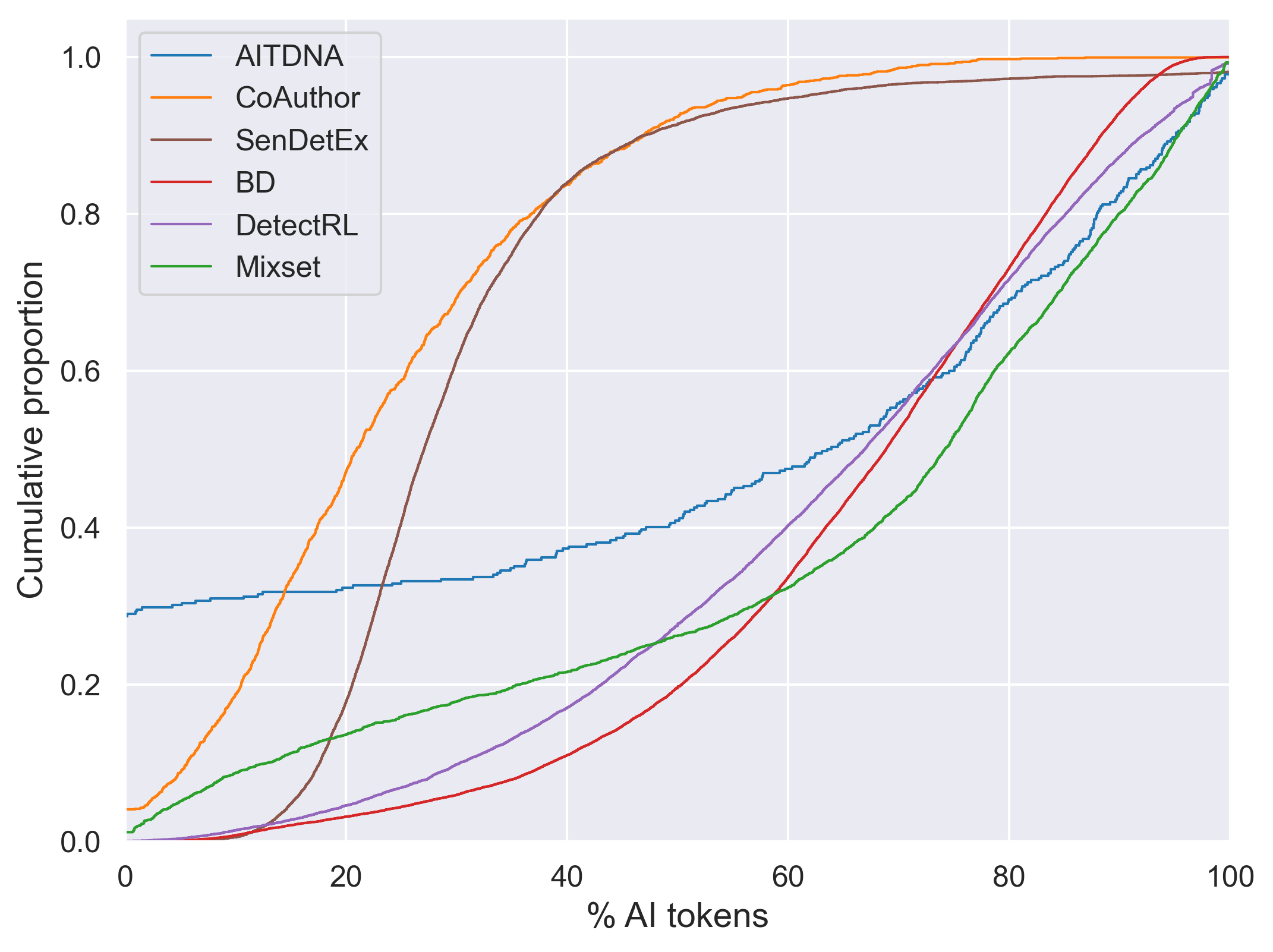}
    \caption{Cumulative proportion of AI token ratios across datasets.}
    \label{fig:ai-token-perc}
\end{figure}

\paragraph{Notion parameters}
We turn to the notion parameters encoded in prior datasets. In particular, genesis-based notions depend on $\tau$, the minimum ratio of AI tokens for a span to be labeled as $\ai$. We analyze the encoded assumptions for $\tau$, focusing on document-level AITD as the most prominent notion.
\Cref{fig:ai-token-perc} reports the cumulative proportion of documents in AITD datasets over the ratio of contained AI tokens, in other words, over $\tau$ on document-level. DetectRL, Mixset, and BD closely align in this regard, with roughly two-thirds of documents containing more than 50\%  AI tokens. On the other end, less than a third of documents in CoAuthor and SenDetEx contain more than 50\% AI tokens. Surprisingly, AITDNA shows a similar cumulative distribution to DetectRL and Mixset but is shifted upwards because of a large portion of mostly human texts. 

We make two key observations: first, although CoAuthor and AITDNA are both notion-agnostic, CoAuthor notably aligns closer to sentence-level AITD assumptions, and AITDNA lies again between dataset assumptions. Second, for DetectRL and Mixset as document-level AITD datasets, $\tau$ de facto lies at 0\%, as all documents in these datasets are labeled $\ai$~ but some have 0\% AI tokens. 
A more realistic choice at $\tau=50\%$ (the majority of content originates from an AI) would lead to notably different datasets for evaluation. Simply put, making $\tau$ explicit is key to ensure alignment with practical notion assumptions and make results on different datasets compatible.

\section{Notions and AITD Evaluation} 
AITD notions define different target labels for the same input text and thus influence detection evaluation. 
We turn to the question: what impact do notions and their parameters have on evaluation?

\subsection{Experimental Setup}
We address this question with three experiments that evaluate detectors under varying notions chosen to reflect plausible real-world needs. 

\paragraph{Detectors.}
To ensure broad coverage of detection approaches, we select multiple best-performing detectors covering both zero-shot and fine-tuned, as well as open-source and commercial detectors. Based on the best models on DetectRL and aligning with \citet{wu-etal-2025-survey}, we use eight open and two proprietary detectors. For the former, we implement Min-K~\citep{shi2024detecting}, Likelihood, Log Rank \cite{gehrmann-etal-2019-gltr}, and Binoculars \cite{abhimanyu-spotting-binoculars-2024} using Gemma-3-1B-it \cite{gemmateam2025gemma3technicalreport} as the surrogate model, which performed best on AITDNA. As top-performing fine-tuned models on the RAID benchmark \cite{dugan-etal-2024-raid}, we use \texttt{modernBERT-ai-detection-raid-mage} (denoted with moBERT) \cite{drayson-etal-2025-machine}, TMR \cite{tmr-ai-text-detector}, and ADAL\footnote{\url{https://huggingface.co/Shushant/ADAL-detector-large}}. Furthermore, we test FastDetectGPT \cite{bao2024fastdetectgpt} using GPT-neo-2.7B for both sampling and scoring as per the original paper. Finally, we employ GPTZero \cite{adam2026gptzerorobustdetectionllmgenerated} and Pangram \cite{emiTechnicalReportPangram2024}, which are proprietary classifiers trained on human, AI, and mixed texts. We map their non-binary outputs (e.g., the 'mixed' label) to $\{\human, \ai\}$ by mapping all labels that are not explicitly $\human$ to $\ai$. Detailed model versions are reported in Appendix \ref{sec:modelconf}.

\paragraph{Metrics.}
We use AUROC, F1-Score, and the False Positive Rate (FPR), as in prior work \cite{wu2024detectrl,dugan-etal-2024-raid}.
In practice, each notion partitions a text $\vd$ into segments $B_{\text{\vd}}$, passed to the detector. We compute the metrics by comparing the detector outputs with the corresponding notion labels $M_{\text{\vd}}$. GPTZero processes the full document and returns sentence-level labels but abstains from prediction for uncertain sentences (e.g., too short sentences), which we treat as an erroneous label by default. Pangram also processes the full document and outputs AI-generated spans, which we map to the respective notion granularity. For example, a segment spanning multiple sentences assigns the same label to each sentence or contained n-gram. Since boundary-level AITD does not adhere to sentence boundaries, mapping the outputs of proprietary detectors is not well-defined, and we do not test them on that notion.

\paragraph{Genesis-based notions.}
We define the notions along typical AI use policies in a university examination setting \cite{wang2024generative}.
We directly derive the surface-level genesis-based AITD notions and set $\tau=0.5$. We select $\beta=5$ for boundary-level AITD slightly higher than related work \cite{zeng2024towards} to account for higher mixing of human-AI text in AITD.
For semantic notions, we define two sets of rules for the content and intent policies. In content-based AITD, we treat AI sentences as violations that directly answer the prompt or contain a key idea or argument. For intent-based AITD, we allow only AI sentences resulting from language polishing requests. Practically, we apply these rules using GPT-5.4-nano in zero-shot mode by first extracting the overall topic and then checking for violations per sentence.\footnote{We report prompts in the supplementary materials.} We verify the classifier's alignment with the intended goals: two co-authors manually annotated 70 randomly sampled AI-generated sentences with their prompts; the chosen subset balances annotation cost and statistical reliability. Annotators showed high agreement, with Krippendorff's alpha reaching $0.80$ for the content policy and $0.88$ for the intent policy. The classifier achieved F1-Scores of $0.89$ (content) and $0.93$ (intent) against the first annotator's labels, and $0.89$ (content) and $0.90$ (intent) against the second annotator's labels, which indicates strong alignment between the classifier and human judgment.

\paragraph{Population-based notions.}
For membership-based AITD, we select the human subset of AITDNA as the reference corpus with $n=2$, which strikes a balance between trivial lexical overlap ($n=1$) and sporadic overlap ($n=3$). This means all texts in the human subset are labeled as fully human since they are in the reference corpus.
Finally, for authorship-ID-based AITD, we tested various n-gram lengths; however, due to the limited number of human texts per author, with the majority of participants contributing only one, no sensible $n$ could be found. This shows a key challenge of authorship-ID-based AITD. We exclude it from the experiments.

\subsection{How Difficult are Notions to Detect?}

\begin{figure*}[t]
    \centering
    \includegraphics[width=\linewidth]{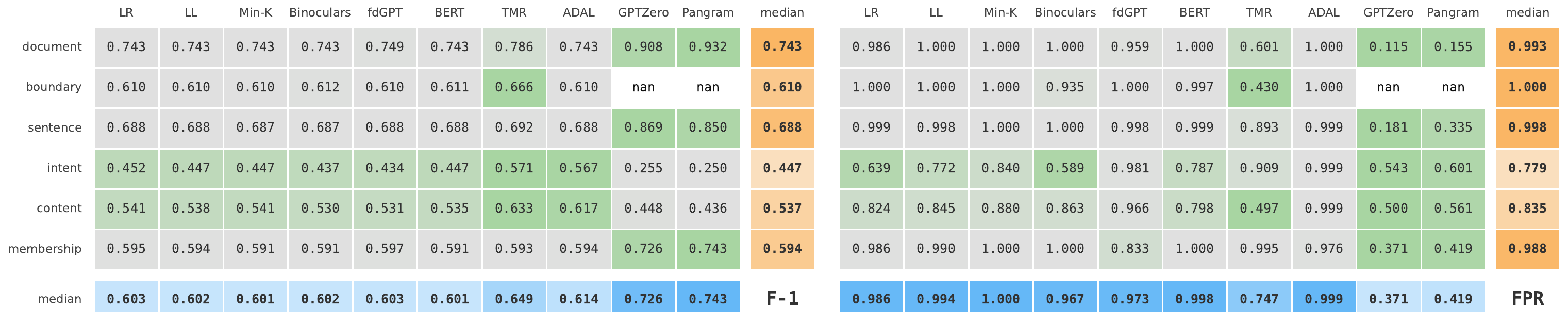}
    \caption{F1-score (left) and FPR (right) per detector and notion in AITDNA. The last column and row report the medians of the detector and notion, respectively. Color gradients indicate better performance per notion and for the medians (higher F1, lower FPR).}
    \label{fig:aitdna-notions-heatmap}
\end{figure*}

In our first experiment, we study the role of notions in evaluation.
By restricting evaluation to AITDNA, we perform controlled experiments testing the impact of different notions on evaluation in the same human-AI co-writing setting. 

We evaluate all detectors on all notions.
\Cref{fig:aitdna-notions-heatmap} shows the F1-score (left) and FPR (right) per detector and notion. We approximate notion difficulty by the median over all detectors per notion. 
Aligned with prior work \cite{jiang-etal-2025-sendetex,saha-feizi-2025-almost}, document-level AITD coincides with the highest detection performance, and detection performance drops with higher granularity. Intent- and content-based AITD are the most difficult to detect for existing detectors, which only consider surface-level features of text and not its content or underlying prompt. Commercial detectors appear to overfit to these features, achieving the lowest performance here while excelling at all others. Surprisingly, they also achieve top performance on membership-based AITD, where most others perform poorly. 
Confirming prior work \cite{dugan-etal-2024-raid}, the FPR on human-AI co-written text is high for most detectors except proprietary ones, indicating over-sensitivity on human-AI co-written text. The FPR of content- and intent-based data compared to sentence-level falls for open detectors, increasing the number of true negatives (labeled $\human$) without increasing the number of false positives. In other words, the permissible samples under our content and intent policies tend to be labeled correctly as $\human$, suggesting that they follow a predictable pattern in terms of token probabilities.

In summary, high granularity leads to lower detection performance, while content- and intent-based AITD remain challenging yet feasible even when relying only on surface features. For policy makers, choosing the right notion means trading off normative standards with detection difficulty.
\subsection{Is there a Best Detector across Datasets?}\label{ssec:best-detector}
AITD datasets impose different notion parameters (\Cref{ssec:assumptions-in-datasets}). We test how detectors perform across datasets when the notion parameters are fixed, focusing on document-level AITD, as it is applicable also to fine-grained AITD datasets. We fix $\tau=0.5$ to detect majority-AI text.

\Cref{tab:detect-compar-by-ds} reports AUROC and F1 of detectors across datasets. We omit Pangram and GPTZero since testing them on all datasets is infeasible with current pricing (over 1500\$ in total). We focus our analysis on AUROC to account for label distribution shifts between datasets (see \Cref{fig:ai-token-perc}).
While aligning $\tau$ enforces consistent notion assumptions, the underlying text sources and the resulting label distribution still lead to variance in performance. There is no consistently best-performing detector across datasets: Log Rank and Likelihood perform best on AITDNA and CoAuthor, respectively, whereas fine-tuned detectors (TMR, followed by moBERT and ADAL) outperform on the remaining four datasets. Since most of these datasets precede the detectors' release (except SenDetEx and moBERT) and may overlap with their training data - for instance, DetectRL and BD overlap with MAGE \cite{li-etal-2024-mage}, which was used to train moBERT - the superior performance may partially be explained by data leakage rather than generalization capabilities. Compared to the majority baseline, hardly any detector other than TMR matches the bar - except for CoAuthor and SenDetEx, where the $\human$ label is the majority, leading to an F1-Score of 0 for $\ai$ as the positive label. Overall, the performance on AI-prevalent datasets is notably higher, suggesting that the high performance may be attributable to a bias towards predicting AI authorship. Our findings only partially agree with the rankings from AITD benchmarks like DetectRL \cite{wu2024detectrl} or RAID \cite{dugan-etal-2024-raid}, which rank fine-tuned detectors like TMR at the top. In our experiments, they perform best on some of the datasets, but are not consistently superior overall. One factor leading to this discrepancy is misaligned notion assumptions, as illustrated by our deviating results on the human subset of DetectRL when fixing $\tau=0.5$. This discrepancy implies the importance of consistent notion parameters. Aligning these for evaluation is key to ensure compatibility of results. Future work on AITD detection should always make these parameters explicit and align them among subsets.

\begin{table*}[t]
\centering
\small
\setlength{\tabcolsep}{4pt}
\begin{tabular}{lcccccccccccc}
\toprule
~
& \multicolumn{2}{c}{\textbf{AITDNA}}
& \multicolumn{2}{c}{\textbf{CoAuthor}}
& \multicolumn{2}{c}{\textbf{SenDetEx}} 
& \multicolumn{2}{c}{\textbf{BD}} 
& \multicolumn{2}{c}{\textbf{DetectRL}}
& \multicolumn{2}{c}{\textbf{Mixset}} \\
\cmidrule(lr){2-3} \cmidrule(lr){4-5} \cmidrule(lr){6-7} \cmidrule(lr){8-9} \cmidrule(lr){10-11} \cmidrule(lr){12-13}
\textbf{Detectors}
& AUROC & $F_1$ 
& AUROC & $F_1$ 
& AUROC & $F_1$ 
& AUROC & $F_1$
& AUROC & $F_1$ 
& AUROC & $F_1$\\
\midrule
majority &  .5 & .743 & .5 & .0 & .5 & .0 & .5 & \underline{.897} & .5 & \underline{.806} & .5 & \underline{.816} \\
\midrule
Log Rank & \textbf{.825} & .739 & .693 & \textbf{.203} & .548 & .18 & .71 & .814 & .573 & .599 & .477 & .765 \\ 
Likelihood & \underline{.815} & .734 & \textbf{.702} & \underline{.195} & .546 & \underline{.185} & .712 & .822 & .559 & .489 & .474 & .796 \\ 
Min K & .732 & .625 & \underline{.689} & .2 & .487 & .145 & .636 & .809 & .543 & .431 & .473 & .772 \\ 
Binoculars & .625 & .637 & .597 & .186 & .428 & .012 & .536 & .729 & .497 & .592 & \underline{.619} & .59 \\ 
fdGPT & .425 & .743 & .45 & .135 & .557 & .180 & .48 & \underline{.879} & .501 & .717 & .406 & \underline{.816} \\ 
moBERT & .754 & \textbf{.828} & .562 & .148 & \underline{.573} & .184 & \textbf{.731} & .722 & \underline{.579} & .45 & .555 & .666 \\ 
TMR & .779 & \underline{.786} & .564 & .138 & \textbf{.645} & \textbf{.253} & \underline{.716} & \textbf{.897} & \textbf{.787} & \textbf{.823} & \textbf{.663} & \textbf{.82}\\
ADAL & .235 & .743 & .478 & .129 & .443 & .169 & .43 & \textbf{.897} & .394 & \underline{.806} & .519 & \underline{.816}\\
\bottomrule
\end{tabular}
\caption{Performance of detectors for document-level AITD across datasets and with $\tau=0.5$ and the best detector per metric and dataset in bold, the second-best underlined. 'majority' is the majority baseline.}
\label{tab:detect-compar-by-ds}
\end{table*}
\subsection{Do Notion Parameters Affect Evaluation?}
Given the previous results, we investigate the role of notion parameters in evaluation, focusing on $\tau$ and document-level AITD and Likelihood as an established AITD method.

\begin{figure}[t]
    \centering
    \includegraphics[width=0.99\linewidth]{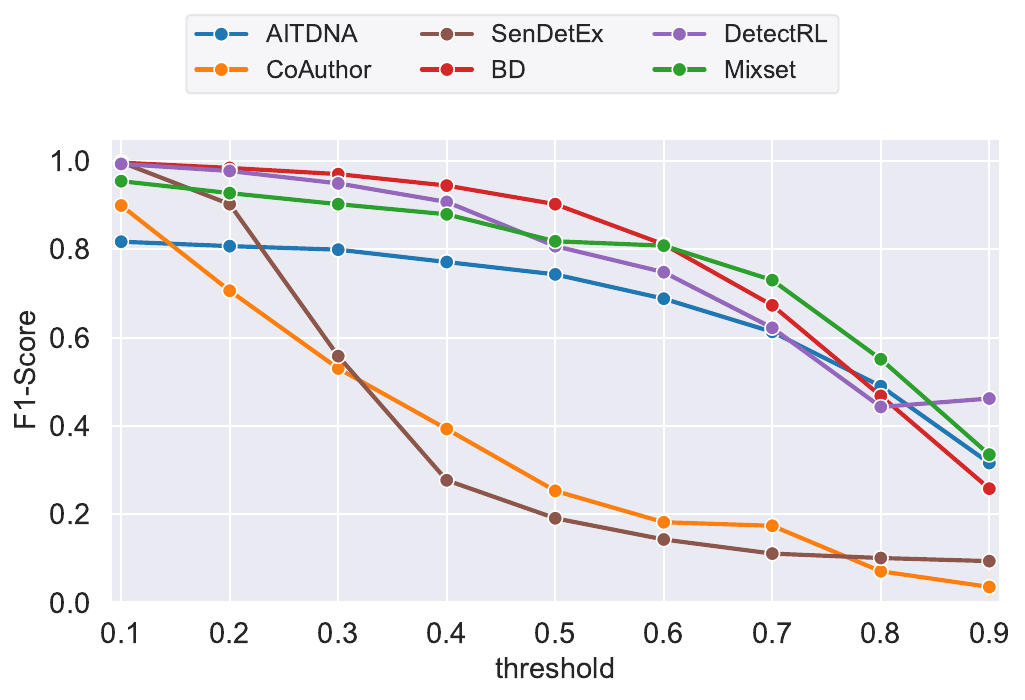}
    \includegraphics[width=0.99\linewidth]{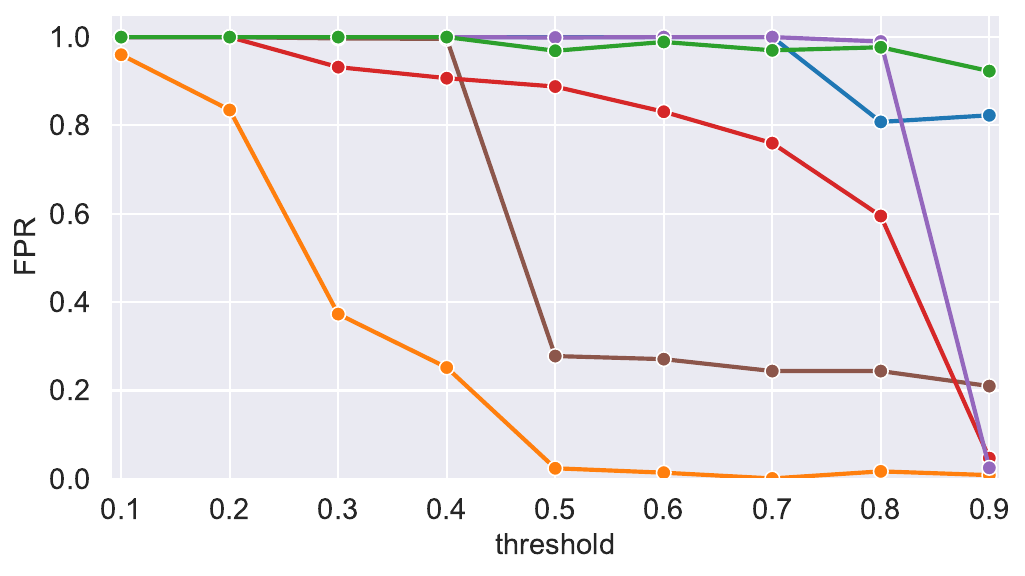}
    \caption{F1-score and FPR of Likelihood per dataset for varying threshold $\tau$.}
    \label{fig:threshold-experiments}
\end{figure}

\Cref{fig:threshold-experiments} reports the F1-score and FPR for varying $\tau$ from 0.1 (documents with $\geq$10\% AI tokens are labeled $\ai$) to 0.9 (with $\geq$90\%).
$\tau$ effectively trades off high F1 with low FPR: for all datasets, both values decrease with growing $\tau$.
If $\tau$ is low, fewer $\human$ texts exist, thus a small decrease in false positives already has a notable impact on FPR. 
The decrease in F1 is linked to the distribution of AI tokens per $\tau$ (see \Cref{fig:ai-token-perc}); since most texts in CoAuthor and SenDetEx have below 40\% AI tokens, F1 and FPR drop before reaching $\tau=0.4$. Although CoAuthor appears as one of the most challenging datasets ( \Cref{tab:detect-compar-by-ds}), this is largely a result of the threshold selection. Across thresholds, it shows among the datasets with the lowest FPR. On the other hand, AITDNA, with seemingly higher detector performance (\Cref{tab:detect-compar-by-ds}), exhibits one of the highest FPR across thresholds.
Since $\tau$ should be chosen to reflect the policies of a use case, many choices are plausible, with $\tau$ typically ranging between 0.1 and 0.5. Even within this range, for all datasets, the F1 varies by at least $0.07$ (on AITDNA) and up to $0.8$ (on SenDetEx). 

Overall, these results show that $\tau$ needs to be carefully chosen relative to the distribution of AI tokens in the target domain. A very small $\tau$ (0.1-0.2) can result in an unacceptably high FPR, while a very large $\tau$ (0.7-0.9) significantly lowers the F1 score. During evaluation and dataset construction, $\tau$ needs to be set explicitly, as performance results otherwise vary strongly. 

\subsection{Are Findings on AITDNA Generalizable?}
AITDNA contains fewer data points than other tested datasets since it relies on time-intensive human labor that cannot be substituted through data synthesis and extends beyond the simplified text interaction setting of CoAuthor. This makes AITDNA useful for many areas of research, including, for instance, human-AI interaction. Nevertheless, the dataset size raises the question: do the findings obtained at the given scale generalize? To investigate this, we estimate the effect of dataset size on the measured detector performance. We focus on CoAuthor, which was collected similarly to AITDNA, subsample it at random 20 times to the size of AITDNA, and compare detector performance on subsamples with that on the full dataset. If the results remain consistent, the AITDNA dataset size is sufficient for the purpose of our study. 
For each subsampled $\dataset_\text{sub}$, we measure AUROC and F1 for all detectors, then compute the mean $\mu$ and standard deviation $\sigma$ per metric across all $\dataset_\text{sub}$. Comparing $\mu$ with the original CoAuthor dataset performance, we compute the absolute bias for both AUROC and F1 in terms of $|\mu-\text{metric}_{\text{orig}}|$. To determine significant deviations, we check if the original measurement falls inside $\mu \pm 2 \sigma$ (about 95\% of the data). \Cref{tab:downsampling} presents the results. For almost all detectors, the subsampled performance remains within the interval, except the Likelihood-F1 combination, which we attribute to the fixed threshold over raw probabilities, making it more sensitive to distribution shifts under subsampling.
Despite this minor difference, the overall ranking of detectors remains insensitive to the dataset size. This underlines the generalizability of our experiments on AITDNA.
\begin{table}[t]
    \centering
    \small
    \setlength{\tabcolsep}{4pt}
    \begin{tabular}{lcccccc}
    \toprule
    & \multicolumn{3}{c}{\textbf{AUROC}} & \multicolumn{3}{c}{\textbf{F1}} \\
        \cmidrule(lr){2-4} \cmidrule(lr){5-7}
        \textbf{Model} & $\mu$ & Bias & $\pm 2\sigma$ & $\mu$ & Bias & $\pm 2\sigma$ \\
        \midrule
        Log Rank & .7 & .007 & True & .261 & .058 & True \\ 
        Likelihood & .704 & .002 & True & .269 & .074 & False \\ 
        Min-K & .701 & .012 & True & .246 & .046 & True \\ 
        Binoculars & .576 & .021 & True & .196 & .01 & True \\ 
        fdGPT & .459 & .009 & True & .138 & .003 & True \\ 
        moBERT & .564 & .002 & True & .147 & .001 & True \\ 
        TMR & .56 & .004 & True & .152 & .014 & True \\ 
        ADAL & .494 & .016 & True & .163 & .034 & True \\ 
        \bottomrule
    \end{tabular}
    \caption{Subsampling experiment results. We measure mean AUROC and F1, absolute bias $|\mu-metric_{true}|$, and whether the metric value of original dataset falls into $\mu \pm 2 \sigma$.}
    \label{tab:downsampling}
\end{table}

\section{Discussion}
Our experiments provide insights into AI-generated text detectors, their datasets, and underlying assumptions. Beyond this diagnostic contribution, they imply practical recommendations to advance AITD research.
First, our findings suggest design guidelines for detectors. Detectors should be developed for specific use cases rather than general applicability. A detector optimized for common AITD benchmarks may fail in contexts such as conferences or university policy settings if it was not designed for them. Second, evaluation should be tailored to the intended use case and hyperparameter choices. For example, the evaluation criteria differ depending on whether the goal is to detect any degree of AI involvement or only high proportions of AI tokens. This choice affects how texts are labeled, regardless of their actual origin. Detecting any AI involvement requires minimizing FPR, whereas detecting high proportions of AI-generated tokens calls for maximizing AUROC or F1. FPR remains a major challenge across the field. Even methods explicitly trained to minimize FPR, such as Pangram, exhibit high FPR, particularly for non-genesis-based notions. Future research should therefore prioritize rigorous FPR evaluation and the development of effective methods for controlling false positives.

\section{Limitations and Future Work} 
Although AITDNA extends related work in terms of interaction types and LLM coverage, we collect the dataset under lab conditions, which influences the way humans write and interact with LLMs. For instance, the typical conversational interaction paradigm of LLMs in popular apps such as ChatGPT is missing, which influences writing outcomes. This points to important future work recording all typical interactions between users and LLMs.
Additionally, our work identifies simplifying and incorrect assumptions in current AITD research. Future work should investigate robust detectors that explicitly account for these assumptions and consistently work across notions. Especially, population-based, content-, and intent-based AITD remain under-explored despite their practical importance.
We proposed authorship-ID-based AITD as a conceptual gap reframing AITD as an authorship identification task relevant to many detection scenarios. However, the collection focus of AITDNA hindered testing this notion practically. Future work should collect dedicated datasets to investigate this notion.

\section{Conclusion}
In this work, we systematically reviewed AITD notions and found that many existing benchmarks make hidden assumptions about what exactly constitutes AI text.
We categorized these assumptions into genesis-based and population-based notions.
Based on this analysis, we introduced new notions that depend on content and authorship identification.
We then introduced AITDNA, a novel, notion-agnostic dataset consisting of human-AI texts that were co-created in a real-world setting.
Due to this, the dataset differs significantly from existing, often synthetic, datasets.
Empirically, we find that there is no detector that consistently outperforms all others across datasets and notions, which shows the importance of awareness of these notions during dataset and detector creation.

\section{Acknowledgments}
This research work has been funded by the German Federal Ministry of Research, Technology and Space and the Hessian Ministry of Higher Education, Research, Science and the Arts within their joint support of the National Research Center for Applied Cybersecurity ATHENE. This work was supported by the Konrad Zuse School of Excellence in Learning and Intelligent Systems (ELIZA\footnote{\url{https://eliza.school/}}) through the DAAD programme Konrad Zuse Schools of Excellence in Artificial Intelligence, sponsored by the Federal Ministry of Research, Technology and Space.
\bibliography{tacl2021}
\bibliographystyle{acl_natbib}

\newpage
\clearpage

\appendix

\section{Notions}
\subsection{Survey Overview}
\Cref{tab:notions} provides the results of the literature survey.

\section{Dataset Creation}
\subsection{Study Interface}
\begin{figure}[ht]
    \centering
    \includegraphics[width=\linewidth]{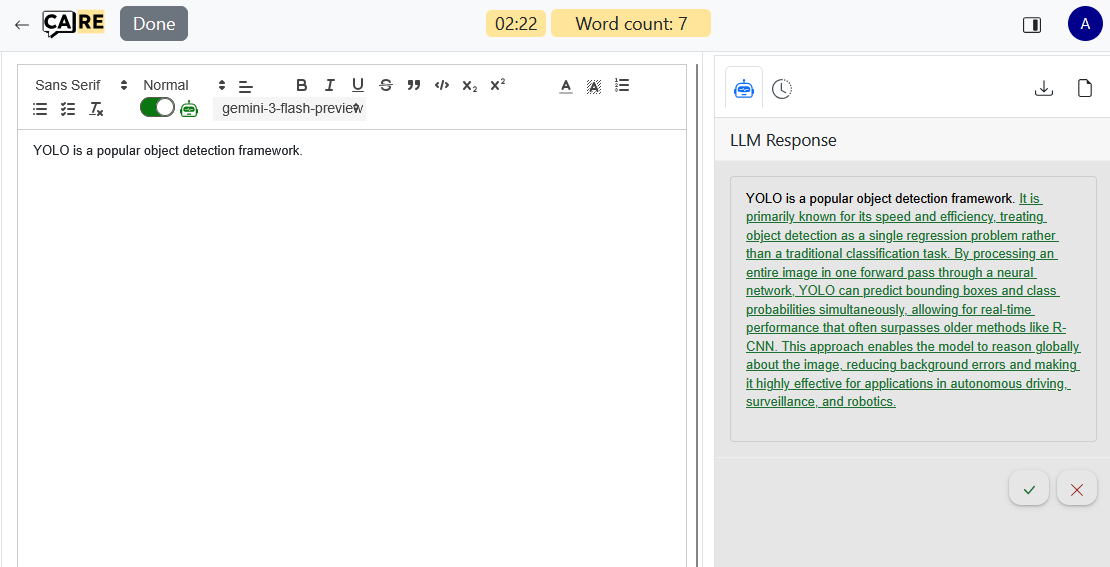}
    \caption{CARE interface for human-AI interaction. On the right side of the interface, an example answer for a continuation query is presented.}
    \label{fig:care-interface}
\end{figure}

\Cref{fig:care-interface} provides an overview of the study interface for participants.
\subsection{Participant Pool}
\begin{figure}
    \centering
    \includegraphics[width=0.9\linewidth]{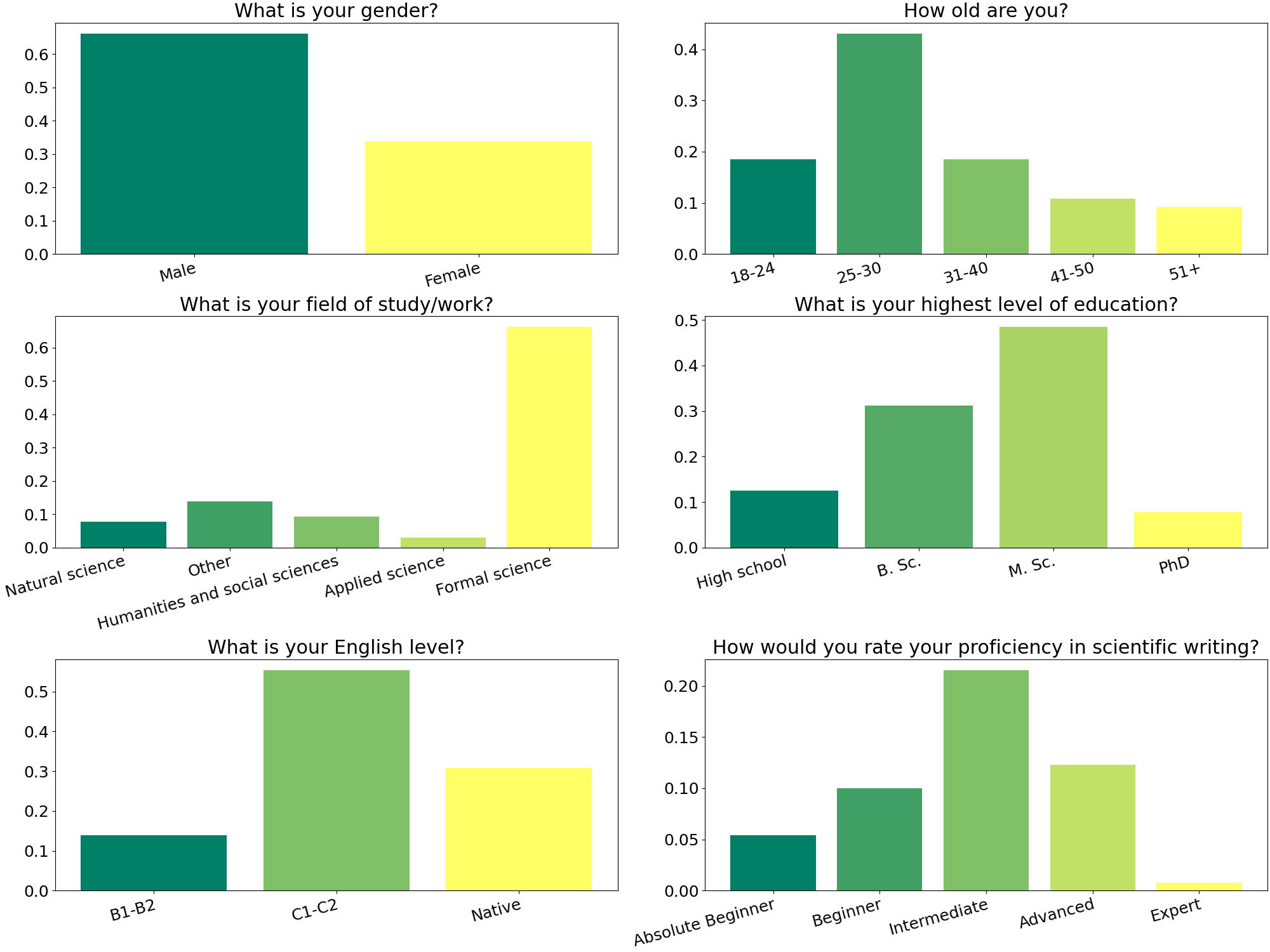}
    \caption{Basic statistics on participant pool computed from the results of the background survey.}
    \label{fig:bg}
\end{figure}

\Cref{fig:bg} displays participant pool statistics, including basic background information.

\section{Experiments}
\subsection{Participant guidelines}
\Cref{fig:study-guidelines} provides condensed guideline text for the participants in the standard setup. \Cref{fig:study-guidelines-scientific} presents writing task descriptions for the scientific setup. Other parts of scientific setup guidelines are identical to the standard setup.
\begin{figure}
    \centering
    \begin{tcolorbox}[fontupper=\tiny]
\textbf{Part 0: Background information survey}\\
Before you start the main part of the study, please fill out this survey on your background information. We ask questions like your age, English level, and field of study.\\\\
\textbf{Part 1: Human-Only Writing}\\
\textbf{Task 1: Explanatory writing Human-Only}\\
In this part, you have 10 minutes to write an explanatory story. You have to write this text on your own, without using LLMs. In the Documents overview, you will see five documents. The first document is \textit{Task 1: Explanatory writing Human-Only}. Click on “Access document” to open it.\\
\textit{Followed by a copy of the task description for the human-AI explanatory setup.}\\

\textbf{Part 2: LLM-powered writing}\\
\textbf{Task 2.0: Warm-up}\\
Open the document \textit{Task 2.0: Warm-up} to start editing. You are now in the document editor where you can type your text and interact with LLMs.\\
There are two available types of interaction: Revision (grammar or stylistic fixes, reformulation) and continuation (text generation like essay wrap up generation). You can send queries in two ways:\\
With the shortcut \textit{CTRL+Q}. To send a revision query, select a part of text to be revised and press the shortcut. A query field will appear where you can type in your custom query. If you don't type anything, the model is prompted to fix the grammar. To send a continuation query, place your cursor in the position to continue the text from, and press the same shortcut. A query field will appear where you can type in your custom query. If you don't type anything, the model is prompted to continue your text.\\
Try using buttons in the bottom-right corner. Select the text you want to revise, or place your cursor in the continuation place. Hover over the buttons to expand the menu. You can use pre-implemented queries or select "Custom query" and enter your own prompt.\\
When the model answer comes back, a new window will appear. You will have the possibility to either accept or reject the proposed text. If you accept it, the text will be copied to the editor. \\
\textbf{Task 2.1: Explanatory writing}\\
In this part, you have 10 minutes to write an explanatory text. You HAVE TO interact with LLMs during the writing.\\
Enter the Document named \textit{Task 2.1: Explanatory Writing}.\\
Explain a concept or a phenomenon that you recently learned and found exciting. Maybe you watched a video about a certain historical period, read a paper with interesting approaches, or learned a new theory at your university. Explain it for general audience, in simple terms: What is the core idea? Where and why is it used? How does it work? What are the benefits and the drawbacks?\\
Examples may be:\\
1. Transformers architecture, used in Large Language Models like GPT-4, is a technology that revolutionized machine text generation and understanding…\\
2. Social engineering is a psychological technique to “hack” people with the purpose to get access to restricted or personal information. …\\
3. There are vast differences in railway systems throughout the world: in terms of reliability, coverage, speed. Japanese trains …\\
Your text should give a clear explanation on the concept/person/time period/technology etc, so that people without deep knowledge can get a good understanding from your text.\\
Approximate expected length is 350-450 words, but it may vary.\\
\textbf{Task 2.2: Creative writing}\\
In this part, you have 10 minutes to write a creative story. You HAVE TO interact with LLMs during the writing.\\
Enter the Document named \textit{Task 2.2: Creative Writing}. Select one of two following topics:\\
\textit{Random topic 1 from the topic list}\\
\textit{Random topic 2 from the topic list}\\
Both questions set up an imaginable scenario where one aspect of our world has changed. You have complete freedom - pick the tone, character, and style that appeal to you. Here are some ideas for inspiration:\\
1. Tell a story from the perspective of an inhabitant of the new world - describe your typical day and how the changes affect your life, work, relationships, etc.\\
2. Write a newspaper-style text that interviews several people from that world.\\
3. Provide an analytical article that analyses the societal, economical, or cultural impact that such changes could have on our society.\\
Approximate expected length is 350-450 words, but it may vary.\\
\textbf{Task 2.3: Argumentative writing}\\
In this part, you have 10 minutes to write an argumentative essay. You HAVE TO interact with the LLM during your writing process.\\
Enter the Document named \textit{Task 2.3: Argumentative Writing}. Select one of two following topics:\\
\textit{Random topic 1 from the topic list}\\
\textit{Random topic 2 from the topic list}\\
Both topics contain several questions. Try to answer them all, or even come up with your own related questions. Your text has to have clear structure and conclusion. Try to find strong arguments to defend your position on the topic, as well as to counter it with opposite arguments. Approximate expected length is 350-450 words, but it may vary.\\
Now you may close the program and move on to the survey!\\\\
\textbf{Part 3: User Experience Survey}\\
After you finished all interactions, please fill out user experience survey. There, we ask about possible problems, limitations, and overall satisfaction. This helps us to understand how we can improve our software.
    \end{tcolorbox}
    \caption{Condensed version of the standard study guidelines provided to the participants. Full version emitted for space constraints.}
    \label{fig:study-guidelines}
\end{figure}

\begin{figure}
    \centering
    \begin{tcolorbox}[fontupper=\tiny]
\textbf{Explanatory writing: Explaining a Research Concept}\\
Explain a scientific concept that you recently learned and found engaging. This could involve describing a paper that proposed an innovative solution to a well-known problem, identifies a gap in the existing literature, or introduces a novel methodology or perspective. E.g.:\\
1. The attention mechanism and its role in today’s LLM architectures \\
2.  Parameter efficient fine-tuning as the key driver for democratization of LLMs \\
3. Inference-time scaling and its tradeoff between efficiency and performance \\
Describe the core idea of this concept, what problem or limitation it addresses, and why it stood out to you. How does it advance the understanding within the field? How does the approach differ from existing methods or theories? How might this idea influence your own research interests or thinking?\\
Your text should give a clear and high-level explanation of the concept, so that people without deep knowledge can get a good understanding from your text. Your goal is to engage and transfer knowledge.\\\\
\textbf{Creative writing: Writing an Impact Statement}\\
Reflect on the implications and consequences of a research idea/direction of your choice. It may be your current work or any topic of interest. Consider the potential impact on the economy, society, and/or culture. Include both positive and negative effects - ethical considerations, unintended consequences, misuse scenarios - and propose ways to mitigate them. Discuss immediate effects, such as outcomes that may occur shortly after publishing a specific paper, as well as indirect or long-term implications, resulting from future research or the broader development of the field. What responsibilities do researchers have in guiding these implications? Note: You should not describe a fully fletched research approach but imagine the “what-if” of any scientific innovation that comes to your mind as impactful. For example:\\
1. If we can successfully automate scholarly peer review, it is the first step towards self-improving artificial intelligence. Researchers will be unburdened from the task. Research progress will be steady. However, selection bias towards certain research will be inevitable. Incentives for gaming the system will be high.\\
2. If we could link each LLM capability to a mechanistic circuit, we would have cracked the LLM black box. For any problem, we could identify the exact underlying causes from the model activations. Yet, this invites new attacks on LLMs in production.\\
Your text has to have a clear structure. Try to think of multiple consequences and perspectives on them. \\\\
\textbf{Argumentative writing: Debating Scientific Claims}\\
Discuss the pros/cons and take a stance on a topic that is debated in your scientific community. You may pick your topic freely (we recommend one from your focus area), but here are some for inspiration:\\
1. Scaling and engineering of existing LLM training technology will lead us towards AGI.\\
2. Differentially private language technology with maximum utility can only be achieved by means of gradient-based privatization strategies.\\
3. Fact checking can and should be fully automatized.\\
State the topic in the beginning and then write your essay. Write a coherent argumentative statement reflecting on the scientific evidence that speaks in favor or against the claim. Target your argumentation towards a reader that is generally knowledgeable in science but not necessarily in your specific field. In the end, take a stance on the topic and link it to the presented arguments.\\
Your text has to have a clear structure and conclusion. Try to find strong pro and con arguments for existing approaches, as well as for your position on the topic.
    \end{tcolorbox}
    \caption{Task descriptions in the scientific setup.}
    \label{fig:study-guidelines-scientific}
\end{figure}

\subsection{Participant Prompts}
 \Cref{fig:topics-creative} and \Cref{fig:topics-argumentative} provide lists of all possible topics for creative and argumentative writing, respectively.
\begin{figure}
    \centering
    \begin{tcolorbox}[fontupper=\tiny]
        \begin{enumerate}[itemsep=0cm, leftmargin=*]
    \item What if no one ever had to work again because robots took care of everything? (\href{https://www.dabblewriter.com/articles/sci-fi-writing-prompts}{source})
    \item What if our memories could be stolen and sold on the black market? (\href{https://www.dabblewriter.com/articles/sci-fi-writing-prompts}{source})
    \item What if we could restore the diminishing populations of endangered species by cloning them? (\href{https://www.dabblewriter.com/articles/sci-fi-writing-prompts}{source})
    \item What if you could test the outcomes of different decisions in virtual reality with 100\% accuracy? (\href{https://www.dabblewriter.com/articles/sci-fi-writing-prompts}{source})
    \item What if there were a world where people spend all their personal time escaping to idyllic VR settings instead of confronting the challenges of real life? (\href{https://www.dabblewriter.com/articles/sci-fi-writing-prompts}{source})
    \item What if the usage of cars for personal needs were banned?
    \item What if people had microchips in their eyes that allowed them to record everything they see?
    \item What if vibe-coding became so popular that every IT company would switch to it?
    \item What if the theory that COVID-19 vaccines contain microchips were true?
    \item What if it were forbidden to sell or buy products containing meat? 
        \end{enumerate}
    \end{tcolorbox}
    \caption{Topics for the creative task. Topics without sources were created by one co-author.}
    \label{fig:topics-creative}
\end{figure}

\begin{figure*}
    \centering
\begin{tcolorbox}[fontupper=\small]
    \begin{enumerate}[leftmargin=*]
        \itemsep0em 
        \item Is it cheating when students use artificial intelligence to help them with their schoolwork? In your opinion, how, if at all, should students be allowed to use AI in school? What do you see as benefits and drawbacks of using AI for doing homework? (\href{https://www.nytimes.com/2025/04/24/learning/are-students-cheating-when-they-use-ai-for-their-schoolwork.html}{source})
        \item Should 16-Year-Olds Be Allowed to Vote? In your opinion, is the current minimum legal voting age of 18 fair and appropriate?  What influence would lowering the threshold to 16 years have on the society? (\href{https://www.nytimes.com/2025/09/04/learning/should-16-year-olds-be-allowed-to-vote.html}{source})
        \item Boys Are Spending More Time Gaming. Is That a Problem? Some say video games are a chief reason boys and young men are struggling. Others say games serve an important role in teens’ lives. What do you think? What can gaming bring to a teen’s life? What other activities, if any, does it take away from? (\href{https://www.nytimes.com/2025/10/07/learning/boys-are-spending-more-time-gaming-is-that-a-problem.html}{source})
        \item Should Schools Ban Student Phones? More and more countries are cracking down on students’ use of cellphones. Are these restrictions fair? Can they work? Do you think that phones interfere with student’s academic learning, the quality of their social interactions or overall engagement in school? (\href{https://www.nytimes.com/2024/09/09/learning/should-schools-ban-student-phones.html}{source})
        \item Is It Ethical for Teachers to Use AI to Grade Papers? Many schools do not allow students to use artificial intelligence to complete their assignments. Should teachers be held to the same standard? Do you think teachers should be able to use the technology at all? If so, in which instances would it be acceptable, and what guidelines should teachers follow when using it? (\href{https://www.nytimes.com/2024/11/13/learning/is-it-ethical-for-teachers-to-use-ai-to-grade-papers.html}{source})
        \item Should Social Media Companies Be Responsible for Fact-Checking Their Sites? A few months ago, Meta, the company that owns Facebook and Instagram, has ended its longstanding fact-checking program. Is that a good idea? What could this mean for users? Do you believe social media companies should be responsible for fact-checking lies, misinformation, disinformation and conspiracy theories on their sites? Why or why not? To what extent does it matter if what we see on social media is true? (\href{https://www.nytimes.com/2025/01/14/learning/should-social-media-companies-be-responsible-for-fact-checking-their-sites.html}{source})
        \item Should Single-Use Vapes Be Banned Everywhere? In an effort to protect young people’s health, England plans to ban disposable vapes next year. Do you think this measure will curb vaping among teens? To what extent do you think the government should have a role in trying to reduce smoking and vaping? Do you think your government should be doing more to discourage people from picking up the habits? Why or why not? What consequences could the ban have for the environment? (\href{https://www.nytimes.com/2024/11/05/learning/should-single-use-vapes-be-banned-everywhere.html}{source})
        \item How Important Is a Free Press to Our Democracy? What role do journalists play in our society - whether they work for big national newspapers, niche podcasts or YouTube channels? What would happen if they couldn’t freely investigate and report the news? For example, what if journalists were barred from informing the public about the actions of powerful people like government officials, corporate executives, and celebrities? How would it affect our democracy? (\href{https://www.nytimes.com/2025/03/27/learning/how-important-is-a-free-press-to-our-democracy-is-it-under-threat.html}{source})
        \item Does Everyone Have a Responsibility to Vote? Is it OK not to vote, or is voting a civic duty? What to you are the most compelling reasons for showing up at the polls? What about those for sitting out? Why do you think so many people don’t vote? What do you think would encourage them to participate more? (\href{https://www.nytimes.com/2024/11/04/learning/does-everyone-have-a-responsibility-to-vote.html}{source})
        \item Workers across industries are concerned about AI coming for their work. Are you worried about AI taking human jobs? Why or why not? Which types of work do you think are most vulnerable to automation? Do you think the fears about AI are overblown? (\href{https://www.nytimes.com/2025/02/07/learning/are-you-worried-about-ai-taking-human-jobs.html}{source})
        \item Should All Children Under 16 Be Barred From Social Media? Australia recently passed a law that does just that. Should other countries do the same? What is your reaction to Australia’s new law? Do you think it is a good idea? Will it be effective? What do you think are the negative effects of social media on young people? If you don’t think a similar law would work, what should be done to address these problems? (\href{https://www.nytimes.com/2024/12/04/learning/should-all-children-under-16-be-barred-from-social-media.html}{source})
        \item Should Grades Be Based on Excellence or Effort? Some people think that too many students today wrongly expect to be rewarded for their efforts rather than the quality of their work. Do you agree? Do you think marks are accurate reflections of students’ learning? What do you think student grades should be based on? How much, if at all, should effort and hard work factor in? (\href{https://www.nytimes.com/2025/01/13/learning/should-grades-be-based-on-excellence-or-effort.html}{source})
    \end{enumerate}
\end{tcolorbox}
\caption{Topics for the argumentative task along with corresponding sources.}
    \label{fig:topics-argumentative}
\end{figure*}

\subsection{LLM Prompt Configurations} \label{sec:promptconf}
\Cref{fig:system-prompt-continuation} and \Cref{fig:system-prompt-revision} provide the system prompts for continuation and revision queries, respectively.\\
\begin{figure}
    \centering
\begin{tcolorbox}[fontupper=\small]
You are a writing assistant. Your role is to continue user text according to their queries.\\
You MUST NOT add greetings, explanations, confirmations, or follow-up offers.
DO NOT repeat user text, just continue it.\\
ALWAYS answer in English!\\\\
Example\\
---------\\\\
User prompt:\\
Write an argument to support my claim.\\\\
Text:\\
In my opinion, the transformers architecture was revolutionary for natural language processing. The reason for that is\\\\
Continuation:\\
the fundamental change of how models understand and generate language. Unlike previous architectures such as RNNs or LSTMs, transformers rely entirely on self-attention mechanisms, allowing them to capture long-range dependencies and contextual relationships within text more effectively.\\\\
Your turn\\
----------\\\\
User prompt:\\
\{\textit{user prompt}\}\\\\
Text:\\
\{\textit{text from CARE}\}\\\\
Continuation:\\
\end{tcolorbox}
\caption{System prompt for continuation queries.}
    \label{fig:system-prompt-continuation}
\end{figure}

\begin{figure}
    \centering
\begin{tcolorbox}[fontupper=\small]
You are a writing assistant. Your role is to modify user text according to their queries.\\
You MUST NOT add greetings, explanations, confirmations, or follow-up offers. 
When responding, start DIRECTLY with the transformed text. Do not include labels, headings, or introductions.\\
ALWAYS answer in English!\\\\
Example\\
---------\\\\
User prompt:\\
Make this text sound better and fix my grammar\\\\
Text:\\
Main advantage of the transformer architecture lays in the ability to handle long-range dependences between different parts of text.\\\\
Revision:\\
The main advantage of the transformer architecture lies in its ability to handle long-range dependencies between different parts of a text.\\\\
Your turn\\
---------\\\\
User prompt:\\
\{\textit{user prompt}\}\\\\
Text:\\
\{\textit{text from CARE}\}\\\\
Revision:\\
\end{tcolorbox}
\caption{System prompt for revision queries.}
    \label{fig:system-prompt-revision}
\end{figure}

\subsection{Detector Configurations} 
\label{sec:modelconf}
For \textit{google/gemma-3-1b-it} and  \textit{EleutherAI/gpt-neo-2.7B}, we use BF16 format, a batch size of 4, and maximum input length of 1024 with truncation. For  \textit{GeorgeDrayson/modernbert-ai-detection-raid-mage} (moBERT), we use a batch size of 4 and maximum input length of 1024 with truncation. Other parameters, such as temperature, were unchanged from the models' default values.




\FloatBarrier
\clearpage
\begin{table*}[h!]
\centering
\renewcommand{\arraystretch}{1.5}
\begin{adjustbox}{angle=90}
\footnotesize
\begin{tabular}{p{2cm} p{1.8cm} p{3.5cm} p{2.5cm} p{3.8cm} p{1cm} p{3cm}} 
\toprule 
\textbf{Datasets} & \textbf{Name} & \textbf{Genesis assumptions} & \textbf{Attacker model} & \textbf{Normative \mbox{standard}} & \textbf{MGT length} & \textbf{Real-world scenario} \\ 
\midrule 

\cite{yu2025cheat,li-etal-2024-mage,verma-etal-2024-ghostbuster} & \textbf{Document-level} & A document is generated \textit{en bloc} by a human or a machine. & Attackers substitute words/phrases in the document. & The document should contain only HGT; there is no benign MGT use. & Several sentences & 
Platforms discover large-scale fake bot networks for propaganda.  \\ 

\cite{zeng2024towards,dugan2023real,wang-etal-2024-m4gt} & \textbf{Boundary-level} & A document is a sequence of up to X passages alternating human and machine text. & Attackers substitute words/phrases in the MGT passages. & There should be no boundaries towards MGT; there is no benign MGT use. & A few sentences & 
News outlets detect large-scale MGT fake news campaigns that remix human-written news stories. \\ 

\cite{ijcai2024p835,jiang-etal-2025-sendetex} & \textbf{Sentence-level} & A document is a sequence of sentences created by a human or machine. & Attackers substitute words/phrases in MGT sentences. & There should be less than Y\% MGT sentences; there exists benign MGT use. & One sentence & 
News readers check what parts of an article are produced by a machine.\\ 

\midrule

\cite{zhang-etal-2024-llm,saha-feizi-2025-almost,saha2026policies} & \textbf{Intent-based} & A document is a sequence of sentences created by a human or machine. & Attackers make changes in MGT related to the input prompt. & There should be less than Y\% MGT sentences violating an intent policy; there exists benign MGT use. & One sentence &An educator checks if passages of an essay were just polished or fully written by an LLM. \\

NEW & \textbf{Content-based} & A document is a sequence of sentences created by a human or machine. & Attackers make semantic changes in MGT sentences. & There should be less than Y\% MGT sentences violating a content policy; there exists benign MGT use. & One sentence & 
Editors check if a research paper has an LLM-generated literature survey.\\ 

\midrule

\cite{triptoShipTheseusCurious2024,koike2025machine} & \textbf{Membership-based} & A document consists of a sequence of n-grams linked to a reference dataset of human n-grams by a similarity metric. & Attackers substitute words in n-grams. & There should be less than Y\% of n-grams below similarity Z; MGT similar to human text is benign MGT use. & n tokens & 
A lawyer checks key contract passages against a corpus of legal documents and jurisdictions.\\ 

NEW & \textbf{Authorship-ID-based} & A document consists of a sequence of n-grams linked to a reference dataset of a specific human's n-grams by a similarity metric. & Attackers make stylistic changes to the n-grams. & There should be less than Y\% of n-grams below similarity Z o the human's reference data; MGT similar to the human's text is benign MGT use. & n tokens & 
An educator checks a student essay submission against prior work of that student.
\\ 

\bottomrule 
\end{tabular}
\end{adjustbox}

\caption{Notions of AITD in the literature structured by their assumptions and normative standards for AI use. The list of literature pointers is non-exhaustive.}
\label{tab:notions}
\end{table*}
\clearpage
\FloatBarrier
\end{document}